# DEEP LEARNING ON SAR IMAGERY: TRANSFER LEARNING VERSUS RANDOMLY INITIALIZED WEIGHTS

*Morteza Karimzadeh[1], Rafael Pires de Lima[1]*

[1]Department of Geography, University of Colorado Boulder

## ABSTRACT

Deploying deep learning on Synthetic Aperture Radar (SAR) data is becoming more common for mapping purposes. One such case is sea ice, which is highly dynamic and rapidly changes as a result of the combined effect of wind, temperature, and ocean currents. Therefore, frequent mapping of sea ice is necessary to ensure safe marine navigation. However, there is a general shortage of expert-labeled data to train deep learning algorithms. Fine-tuning a pre-trained model on SAR imagery is a potential solution. In this paper, we compare the performance of deep learning models trained from scratch using randomly initialized weights against pre-trained models that we fine-tune for this purpose. Our results show that pre-trained models lead to better results, especially on test samples from the melt season.

## 1. INTRODUCTION

In recent years, various architectures of deep learning have been developed for Synthetic Aperture Radar (SAR) imagery in application domains spanning environmental monitoring and change detection. One such case is sea ice mapping. SAR is the primary data source for mapping sea ice, as multiple C-band SAR sensors including Sentinel-1 and RADARSAT-2 have polar coverage, and can acquire images regardless of cloud cover or light conditions. Sea ice undergoes constant and rapid changes due to the combined influence of wind, temperature, and ocean currents. Hence, frequent mapping of sea ice is essential to ensure maritime safety. Currently, sea ice mapping is primarily performed by national ice centers of countries having interests in the Arctic and Antarctic regions, as automated mapping of sea ice using SAR imagery still remains a challenge, especially during the melt season, when surface melt masks the underlying ice surface, resulting in mistaking ice for open water.

Deploying deep learning on SAR imagery is challenging for several other reasons as well, including (a) the systematic TOPSAR noise (banding and scalloping) in the Extra Wide (EW) mode (which is the sole mode of acquisition over open oceans and polar regions), (b) ambiguous volume scattering patterns of sea ice types with different thickness, (c) similar backscatter patterns of smooth dark young ice and calm water, making the discrimination of water and ice challenging. Researchers have been experimenting to establish optimal configuration and training strategies for deep learning models that best tackle these challenges. One important area, less explored systematically, is fine-tuning image segmentation models on SAR imagery using models pre-trained on natural RGB imagery [1], [2]. Given the inherent differences of SAR and optical imagery, as well as differences of remote sensing and generic/natural (fashion, and animal) targets, it is unclear what impact starting with pre-trained weights would have on the results of segmentation.

In this paper, we analyze the performance of deep learning-based image segmentation models on SAR imagery using two different training strategies: one using transfer learning, fine-tuning pre-trained ImageNet weights on SAR imagery, and the other strategy using randomly initialized weights. We use a publicly available benchmark dataset for this purpose, and test the model performance on held-out test scenes, one during the melt season and one during the freeze up season. We analyze the results using both performance metrics, as well as visual inspection of classification error for each set up.

## 2. DATA AND MODEL

We use the Extreme Earth v2 dataset [3], which includes high-resolution ice charts over the East Coast of Greenland aligned with twelve Sentinel-1 images acquired in EW mode, with each image having a spatial footprint 400 x 400 km. The twelve images were acquired roughly one month apart throughout 2018. The polygon labels are interpretations of expert sea ice analysts using SAR as primary source, as well as other data sources used in conjunction with domain knowledge of the region.

We use the labels to train semantic segmentation models for the separation of ice and water. We hold out image and label pairs acquired in January and July (two out of twelve) for testing the performance of the model, with January representing the freeze up season conditions, and July for the melt season, which as mentioned above, is more challenging for deep learning models.

For validation during training using non-overlapping images, we clip half of the entire February, June, August, and December images, and assign them to the validation (i.e., development) set. The training samples are generated by the extraction of 100 randomly placed patches of size 80 km, equivalent to 1000 x 1000 pixels using images with 80 x 80 m pixel size. Since our models are fully convolutional, we generate output for test and validation images using a single pass on the entire scene. Our model architecture uses the first three blocks of ResNet18 [4] as encoder, and a decoder based on the Atrous Spatial Pyramid Pooling (ASPP) module [5], resulting in a total of 4 M trainable parameters.

The model takes as input the horizontal emit, horizontal receive (HH) and horizontal emit, vertical receive (HV) polarization values of SAR, in addition to the incidence angle from Sentinel-1 EW mode, and rasterized ice and water polygons from the Extreme Earth dataset as labels. Raster labels are binary, with one class representing water and another representing ice. We use a batch size of 32, Adam optimizer [6] with a learning rate starting at 1e-5 to train the models. We decrease the learning rate by a factor of 10 when the validation loss does not decrease in five epochs, to a minimum of 1e-8. The models stop training when the validation loss does not decrease in 20 epochs. We save the models' weights with the smallest validation loss for testing. These hyperparameters are kept the same for all models trained.

### 3. EXPERIMENTS

We perform two sets of experiments, with three runs for each to average the performance metrics over the stochastic nature of gradient descent optimization. First, we initialize the entire model with random weights using PyTorch's [7] default parameters, hereinafter "randomly initialized models". Second, we initialize the decoder with random weights, but the encoder is initialized with ImageNet [8] weights. The weights of the encoder and decoder are updated during training. We call these "pre-trained" models.

### 4. RESULTS

Table 1 shows the average for resulting metrics for the experiments across three runs for each setup. Our results show that pre-trained models have better performance metrics than randomly initialized models on average for the melt season test scene (i.e., July). Specifically, weighted F1 increases by 0.06 to 0.98 and weighted IOU increases by 0.11 to 0.95, which is a considerable improvement.

As for the July scene, there are noteworthy observations (Fig. 1): pre-trained models are more robust and classify ice under banding noise, and better classify water under windy conditions. Randomly initialized models are thrown off by ruffled water as well as banding noise over areas of low backscatter such as dark (younger) first year ice.

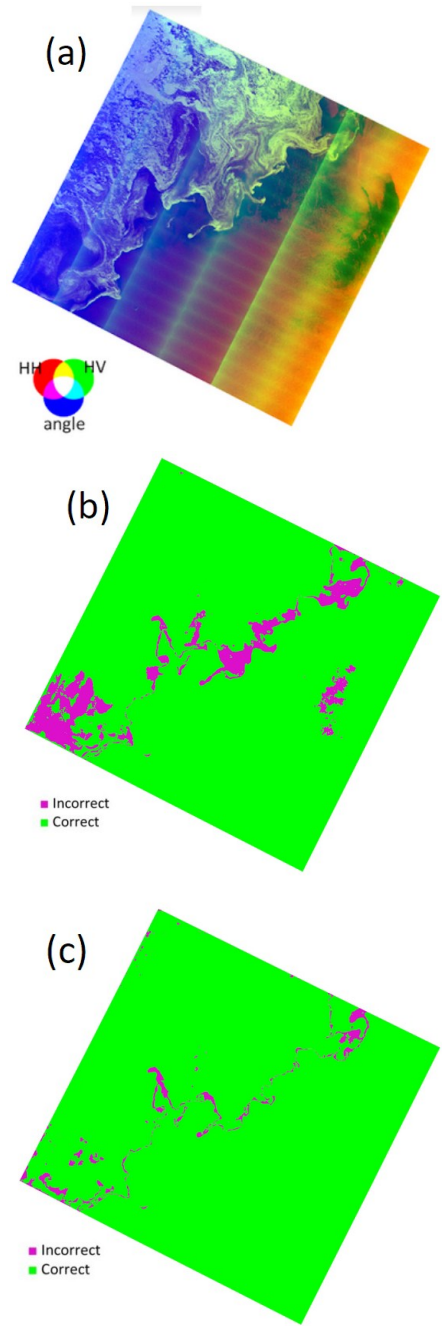

**Fig. 1.** (a) SAR image acquired in July from the Extreme Earth V2 dataset (b) randomly initialized model misclassification error in purple for the same image, (c) pre-trained model classification error map. Fine-tuning a pre-trained model has led to much better results during the melt season.

TABLE 1: Performance metrics comparison for the two setups, averaged for three training runs to minimize stochasticity.

|  |  | Average F1 | Micro avg IOU | Macro avg IOU | Weighted IOU |
|---|---|---|---|---|---|
| January test scene | Randomly initialized | 0.98 | 0.96 | 0.96 | 0.96 |
|  | Pre-trained | 0.97 | 0.95 | 0.95 | 0.95 |
| July test scene | Randomly initialized | 0.92 | 0.85 | 0.85 | 0.85 |
|  | Pre-trained | 0.98 | 0.96 | 0.96 | 0.96 |

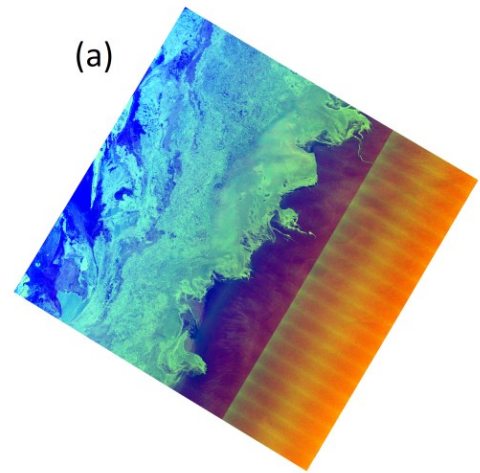

Looking closer at the confusion matrix for the July test scene, we observe that there are major improvements in identifying sea ice when fine-tuning a pre-trained model. When using randomly initialized weights, 15% of actual sea ice pixels are mistakenly classified as water, which can lead to potentially risky outcomes for generating navigational ice charts.

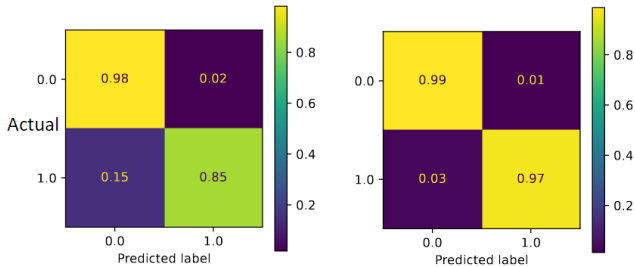

**Fig. 2.** Confusion matrix for the July (melt conditions) test scene, for the model with randomly initialized weights (left) Confusion matrix for the model fine-tuned from with pre-trained weights (right), which shows much better performance on the sea ice class. In the legend, class 0 represents water, and class 1 is sea ice.

While using pre-trained models has a clear advantage on the melt-season test scene, the results on the January test scene are not as conclusive, and in fact, show potentially opposite effects in performance. Metrics are similar on the January (freeze up) test scene for both models, with F1 around 0.97 and IOU approximately 0.95, and a 0.1 decrease in performance for both metrics when fine-tuning pre-trained models compared to randomly initialized weights. Figure 3 shows misclassification errors for the January test scene, with both models having roughly similar results, with the model with randomly initialized weights slightly more successful in classifying sea ice along the edge (Fig. 3 And Fig. 4), however, the model with pre-trained weights shows slightly better results for classifying water.

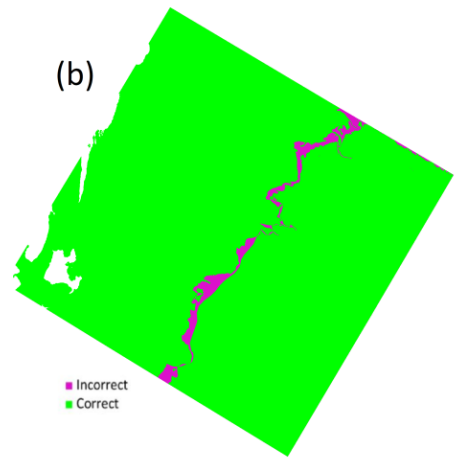

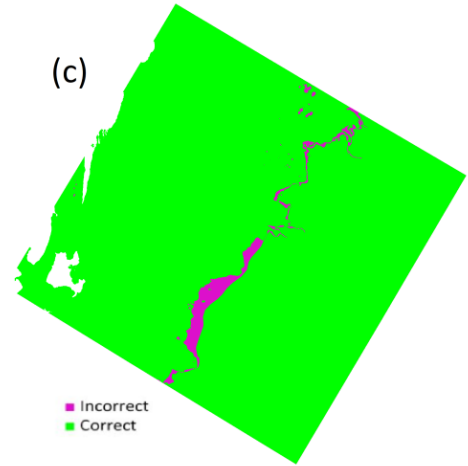

**Fig. 3.** (a) SAR image acquired in January from the Extreme Earth V2 dataset, (b) pre-trained model classification error map, (c) randomly initialized model misclassification error in purple for the same image. Both models perform roughly similarly overall on both classes, however, the pre-trained

model performs slightly better in identifying pixels of the sea ice class.

It is worth noting that pre-trained models tended to train faster too, unsurprisingly: the number of epochs for pre-trained models to stop training was 28, 29, 39 for the three experiments, against 32, 36, 45 epochs for models with randomly initialized weights.

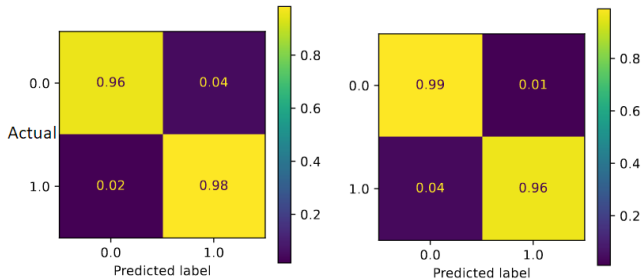

**Fig. 4.** Confusion matrix for the January (freeze up conditions) test scene, for the model with randomly initialized weights (left) and the model fine-tuned with pre-trained weights (right).

## 5. CONCLUSION AND FUTURE WORK

In this study, we compared the performance of fine-tuning deep-learning-based segmentation models pre-trained on natural images against models trained on randomly initialized weights for the purpose of sea ice mapping. Our results highlight the potential of fine-tuning models originally pre-trained on generic images for use with SAR imagery in mapping sea ice, leading to better performance and usually fewer epochs to converge. The results show clear improvement for samples collected during the melt season, when sea ice mapping is commonly more challenging due to similarities in signal of open water, melt ponds, and generally, surface melt. However, the results for samples collected during the freeze up season are not conclusive, with only a slight advantage for the models initialized with random weights for classifying sea ice, and slight advantage for pre-trained models for classifying water.

Future research on larger datasets is needed to further explore the effects of pre-trained weights on model output. Additionally, tasks such as sea ice type classification, concentration estimation, and floe size estimation require similar analyses. Research into using different model sizes (layers) and the specific pre-trained weights (coming from different generics datasets) can also help pave the way for more efficient model design and implementation with fewer training samples for sea ice mapping, and remote sensing with SAR in general.

## 6. ACKNOWLEDGEMENT

This material is based upon work supported by the National Science Foundation under Grant No. 2026962. We thank the Extreme Earth project and MET Norway for making the ExtremeEarth dataset available to the sea ice community. The code used for this research is available at https://github.com/geohai/sea-ice-segment.